\documentclass[sigconf]{acmart}

\usepackage{colortbl}
\usepackage{tabularx}
\usepackage{float}

\AtBeginDocument{%
  \providecommand\BibTeX{{%
    \normalfont B\kern-0.5em{\scshape i\kern-0.25em b}\kern-0.8em\TeX}}}

\setcopyright{rightsretained}
\copyrightyear{2024}
\acmYear{2024}
\acmConference[Genaiecom '24]{Proceedings of the first workshop on Generative AI for E-Commerce}{October 25, 2024}{Boise, ID}
\acmBooktitle{Proceedings of the first workshop on Generative AI for E-Commerce 2024, October 25, 2024}
\acmYear{2024}
\acmMonth{10}
\acmDOI{}
\acmISBN{}
\acmPrice{15.00}

\graphicspath{{./images/}}

\begin{document}

\title{Hierarchical Knowledge Graph Construction from Images for Scalable E-Commerce}


\author{Zhantao Yang}
\affiliation{%
  \institution{Carnegie Mellon University}
  \city{Pittsburgh}
  \country{USA}}
\email{zhantaoy@andrew.cmu.edu}

\author{Han Zhang}
\affiliation{%
  \institution{Carnegie Mellon University}
  \city{Pittsburgh}
  \country{USA}}
\email{hanz3@andrew.cmu.edu}

\author{Fangyi Chen}
\affiliation{%
  \institution{Carnegie Mellon University}
  \city{Pittsburgh}
  \country{USA}}
\email{fangyic@andrew.cmu.edu}

\author{Anudeepsekhar Bolimera}
\affiliation{%
  \institution{Carnegie Mellon University}
  \city{Pittsburgh}
  \country{USA}}
\email{abolimer@andrew.cmu.edu}

\author{Marios Savvides}
\affiliation{%
  \institution{Carnegie Mellon University}
  \city{Pittsburgh}
  \country{USA}}
\email{	marioss@andrew.cmu.edu}


\begin{abstract}
  Knowledge Graph (KG) is playing an increasingly important role in various AI systems. For e-commerce, an efficient and low-cost automated knowledge graph construction method is the foundation of enabling various successful downstream applications. In this paper, we propose a novel method for constructing structured product knowledge graphs from raw product images. The method cooperatively leverages recent advances in the vision-language model (VLM) and large language model (LLM), fully automating the process and allowing timely graph updates. We also present a human-annotated e-commerce product dataset for benchmarking product property extraction in knowledge graph construction. Our method outperforms our baseline in all metrics and evaluated properties, demonstrating its effectiveness and bright usage potential.
\end{abstract}




\maketitle

\begin{figure*}[!htb]
    \centering
    \includegraphics[width=1.0\textwidth]{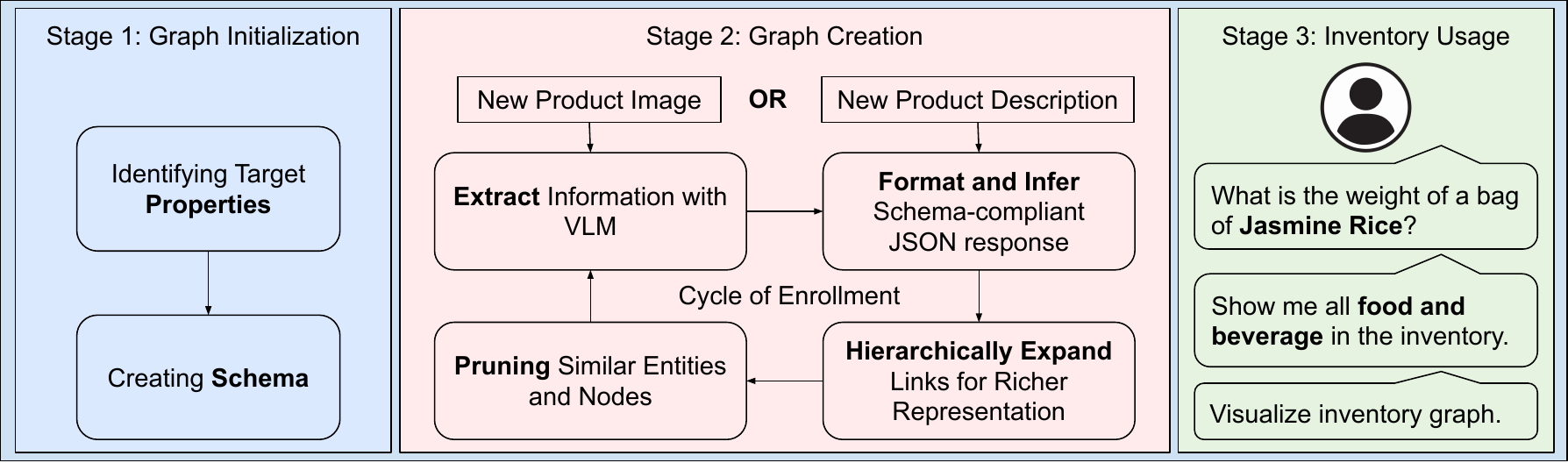}
    \Description{An overview of the method. Stage 1: An empty graph is first initialized with target properties and corresponding data types. Stage2: for each product}
    \caption{ 
    Method Overview. Stage 1: An empty graph is first initialized with target properties and corresponding data types. Stage2: for each product, information is extracted with VLMs, organized and improved by LLM.
    }
    \label{fig:fig1}
\end{figure*}

\section{Introduction}

Knowledge graphs (KG), directed graphs representing information and relationships between entities, are commonly used for efficient information processing. In the domain of e-commerce, KGs play a crucial role in scaling up both inventory management and customer service by leveraging various applications \cite{https://arxiv.org/pdf/2303.13948, https://www.mdpi.com/2078-2489/13/4/161}, including recommendation systems \cite{https://link.springer.com/content/pdf/10.1007/s42979-023-02149-6.pdf, https://arxiv.org/pdf/1909.12807, https://arxiv.org/pdf/2003.00911, https://www.sciencedirect.com/science/article/pii/S0957417420305881, https://pdfs.semanticscholar.org/fe4f/f1a12bf198d2c0fdcfd3757f8ec204355a75.pdf}, question answering service \cite{https://ceur-ws.org/Vol-2722/nliwod2020-paper-3.pdf, https://www.mdpi.com/2079-9292/12/15/3363, https://arxiv.org/pdf/1912.05728}, information and intention discovery \cite{https://arxiv.org/pdf/2211.08316}, and knowledge completion \cite{https://arxiv.org/pdf/2310.20588}.

Recent studies \cite{https://arxiv.org/pdf/2406.02962, https://arxiv.org/pdf/2305.13168, https://arxiv.org/pdf/2404.19146, https://arxiv.org/pdf/2404.03868} show improvements extracting information from documents and texts for knowledge graph construction. However, in practice, due to the rapid changes in the fields of e-commerce, informative text descriptions of products are often expensive and time-consuming to acquire through human labeling. In contrast, raw images of products \cite{https://arxiv.org/pdf/1901.07249, https://arxiv.org/pdf/2006.12634v7} are widely available yet under-explored as sources of automated knowledge graph construction. 

In this work, we explore how to establish an automatic process that directly uses product images as the primary sources to construct complex knowledge graphs. Without human-in-the-loop, the process of populating knowledge graphs particularly benefits the fast-paced e-commerce sector, where product catalogs are constantly evolving and expanding, so that timely update is achieved in such an environment. Moreover, product images contain essential information that is language-agnostic, semantic-rich, and involves subtle visual cues, ensuring accurate product representations toward multilingual and multicultural e-commerce platforms.


Despite the advantages, establishing an automatic process for KG with product images is a non-trivial work and faces many challenges. 
\textbf{Firstly}, unlike documents and texts which directly include the entities, properties, and relationships, product images are complex and may contain distractions. Extracting useful information thus requires sophisticated image understanding.  \textbf{Secondly}, not all relevant information for KG construction is directly visible or explicitly stated in the product image itself. For example, categorizing a chocolate image into candy requires the ability to reason based on common knowledge and contextual understanding. \textbf{Thirdly}, unconstrained triple generation may not fully capture the hierarchical nature of product properties in e-commerce. For instance, the category property can have a hierarchical relationship, "chocolate" falls under "candy," which falls under "food". Many previous works construct KGs by either generating \cite{https://arxiv.org/pdf/2305.13168, https://arxiv.org/pdf/2406.02962, https://ersj.eu/journal/3395} or completing \cite{https://arxiv.org/pdf/2305.09858, https://arxiv.org/pdf/2104.07650} triples without additional constraints, which could result in graphs lacking depth and diversity if directly applied to e-commerce products.

To address these limitations, we propose a novel method that is equipped with the recent advances in vision-language models (VLMs) and large language models (LLMs), enabling hierarchical knowledge graph generation given any number of product images. The graph follows a schema containing both properties and their corresponding data types. We start with instruction-tuned vision language model (InternVL2-8B 
\cite{https://arxiv.org/pdf/2312.14238, https://arxiv.org/pdf/2404.16821}) to extract detailed information from product images. Then, we use schema-augmented multi-turn conversation to ensure the description contains more diverse and detailed attributes and relationships. We further use the newest state-of-the-art large language model, Llama3.1-70B \cite{https://arxiv.org/pdf/2407.21783}, to reason and infer KG relevant properties not found in the image, and hierarchically expand the existing links. In the process, SGLang \cite{https://arxiv.org/pdf/2312.07104} is selected to generate all LLM responses in strictly reliable structured formats, ensuring the output graph follows the schema. Finally, we design programmatical merge to reduce redundancy among similar entities.

Besides the proposed method, we introduce a human-annotated e-commerce product image dataset. It consists of 105 images, each containing a product image, 4 categorical properties, 1 numerical property, and JSON structured metadata.  We benchmark our method on the dataset and release the dataset to the community for further research. 

Our contribution is in three folds:

\begin{itemize}
  \item To the best of our knowledge, we are the first to propose a novel fully automatic e-commerce method to generate knowledge graphs from only product images.
  \item We present a small e-commerce product image dataset for benchmarking the task.
  \item Our method outperforms the baseline method modified from previous works on multiple metrics.
\end{itemize}



\section{Related Work}

\subsection{Text Extraction from Images}
Image understanding has made significant strides in recent years, particularly in the domain of text extraction from images. Image captioning models aim to generate natural language descriptions of images. Image captioning models like LEMON \cite{https://arxiv.org/pdf/2111.12233} are often based on encoder-decoder architectures. The encoder projects images into latent space, and a language decoder decodes semantic information into text descriptions. However, their outputs are often general and lack the specificity required in e-commerce applications. Image tagging models like RAM \cite{https://arxiv.org/pdf/2306.03514} take a step forward, focusing on identifying and labeling specific objects within an image, providing more fine-grained information, but still lacking the capability of identifying texts and concepts. The emergence of multimodal LLMs \cite{https://arxiv.org/pdf/2306.13549, https://arxiv.org/pdf/2401.13601}, especially instruction-tuned large vision language models \cite{https://arxiv.org/pdf/2304.08485}, provides a significant advancement in image understanding. These VLMs can output detailed descriptions based on user prompts, making it possible to align generated texts with desired properties. In this work, we use InternVL2-8B \cite{https://arxiv.org/pdf/2312.14238, https://arxiv.org/pdf/2404.16821}, a robust open-source VLM as our image description extractor. While the model provides detailed descriptions, this information is unstructured and may not contain all the required information for KG construction.

\subsection{Knowledge Graph Construction}
Knowledge graph construction aims to convert less organized raw data into more programmatically processable structured graphs. Recent advancements in deep learning and natural language processing (NLP) \cite{https://arxiv.org/pdf/1706.03762, https://arxiv.org/pdf/2005.14165, https://arxiv.org/pdf/2303.08774, https://arxiv.org/pdf/2203.02155} have significantly enabled studies in information extraction and knowledge graph construction, leading to improved data management and utilization. Due to the advancements in embedding training, some studies use embeddings to represent and discover complex structures in KGs. ComplEX \cite{https://arxiv.org/pdf/1606.06357} uses complex-valued embeddings to perform link prediction at a linear time and space complexity. KoPA \cite{https://arxiv.org/pdf/2310.06671} performs triple classification task by training an LLM adapter for injecting structural embedding. However, these works are limited to individual subtasks of completing a KG. Consequently, many works have shifted their attention to constructing knowledge graphs from documents \cite{https://arxiv.org/pdf/2406.02962, https://arxiv.org/pdf/2308.11730, https://arxiv.org/pdf/2404.09416} and texts \cite{https://arxiv.org/pdf/2305.13168, https://arxiv.org/pdf/2404.19146, https://arxiv.org/pdf/2404.03868}. These recent approaches have leveraged advancements in transformer \cite{https://arxiv.org/pdf/1706.03762} architecture based large language models (LLMs) trained on internet-scale datasets \cite{https://jmlr.org/papers/volume21/20-074/20-074.pdf, https://d4mucfpksywv.cloudfront.net/better-language-models/language-models.pdf} to enhance or construct knowledge graphs. For example, TKGCon \cite{https://arxiv.org/pdf/2404.19146} uses GPT-4 \cite{https://arxiv.org/pdf/2303.08774} to generate theme-related entities and relations from a theme-specific corpus to form KGs. However, few works are leveraging raw images for KG construction. In this work, we use Llama3.1 \cite{https://arxiv.org/pdf/2407.21783}, in collaboration with a VLM to generate high-quality and diverse knowledge graphs for e-commerce product inventory.

\section{Constrained Hierarchical Knowledge Graph Generation}
In this section, we will show an overview of our Knowledge Graph construction method.

\subsection{Method Overview}
The knowledge graph construction can be roughly divided into two core stages. An abstract visualization of our method can be found in Figure \ref{fig:fig1}.

\begin{itemize}
  \item \textbf{Graph Initialization.} When creating a new knowledge graph for an inventory or an e-commerce system, two components are used for initializing and preparing the knowledge graph: \textbf{identifying target properties} and \textbf{creating schema}. An empty inventory knowledge graph will be created once initialization is finished.
  \item \textbf{Cycle of Enrollment.} An e-commerce inventory can grow continuously as more products are added. Therefore, our proposed method treats each product as the fundamental unit of graph creation. Our method cycles through four sequential steps for each product. A product-centric knowledge graph will be generated with four steps: \textbf{Extracting}, \textbf{Formatting and Inferring}, \textbf{Hierarchy Expansion}, and \textbf{Graph Pruning}. Each product knowledge graph will be added to the previous inventory KG.
  \item \textbf{Inventory Usage} Once the knowledge graph is initialized, it can be loaded into a graph database and used in various downstream applications.
\end{itemize}

\section{Introduction}

\subsection{Graph Initialization}

\textbf{Identifying Target Properties.} The first step in the initialization phase is to identify and select the properties that will serve as the foundation of the knowledge graph. Not all entities and relationships are essential when utilizing a knowledge graph for e-commerce. The process of identifying target properties aims to determine the most relevant and valuable attributes for products in the inventory. This can be done automatically, manually, or semi-automatically. In automatic mode, the pipeline prompts an instruction-tuned LLM \cite{https://arxiv.org/pdf/2203.02155, https://arxiv.org/pdf/2407.21783} to list the most important properties when describing an e-commerce product. Alternatively, a person can designate the properties that are important for their system.

\textbf{Creating Schema.} While the properties to be generated have been defined, it is important to standardize the structure and format of different products. The schema defines the data types of each property, which serves as a blueprint for each new product subgraph. By enforcing rules and constraints, the schema helps reduce the introduction of redundant or conflicting data when new products are added, which provides better scalability and downstream task efficiency. Enforcing data types also acts as a fail-safe, preventing LLMs from generating invalid information. In this work, we use a prompted autoregressive LLM to find the data type $t$ that maximizes the predicted probability for each given property $x$:
\begin{equation}
    t' = \text{argmax}_{t' \in \{int,float,str,choices\}} P(t|x)
\end{equation}

If the data type of a property is identified as int or float, a unit of measurement is similarly predicted with an autoregressive LLM. The model predicts the next token following the prompt "\{property\} of a product could be 5 ". If the data type is identified as choices, LLM is prompted to generate diverse distinct choices that can generalize to most products, with an additional "Other" choice added.

Following the above procedure, a complete schema can be created. A product subgraph takes the product name as the root node, all edges point to properties starting from the product root node. By default, we use the schema generated fully automatically:
\begin{itemize}
    \item \textbf{Product Name}: \textit{string}
    \item \textbf{Category}: \textit{choices} [Electronics, Fashion, Home and Kitchen, Beauty and Personal Care, Food and Beverages, Sports and Outdoors, Baby and Kids Products, Health and Wellness, Automotive, Arts and Crafts, Pet Products, Office and School Supplies, Industrial and Scientific, Musical Instruments, Toys and Games, Others]
    \item \textbf{Brand}: \textit{string}
    \item \textbf{Price}: \textit{float} (USD)
    \item \textbf{Primary Package Color}: \textit{choices} [White, Black, Gray, Beige, Brown, Tan, Green, Red, Blue, Yellow, Light Blue, Pink, Baby Blue, Mint Green, Silver, Gold, Copper, Purple, Orange, Turquoise, Others]
    \item \textbf{Package Material}: \textit{choices} [Plastic, Paper, Cardboard, Glass, Metal, Wood, Fabric, Foam, Bamboo, Bioplastic, Molded Pulp, Corrugated, Others]
    \item \textbf{Package Shape}: \textit{choices} [Rectangular, Cylindrical, Spherical, Oval, Triangular, Irregular, Flat, Tubular, Conical, Geometric, Others]
    \item \textbf{Weight}: \textit{float} (kg)
\end{itemize}

\begin{table*}[htbp]
\centering
\caption{Comparison of our method against the baseline on various properties. Accuracy is reported for categorical properties, while accuracy@threshold is used for numerical properties. All results shown in percentage (\%)}
\label{tab:results}
\renewcommand{\arraystretch}{0.9} 
\setlength{\tabcolsep}{4pt} 
\begin{tabularx}{\textwidth}{l|*{6}{>{\centering\arraybackslash}X}}
\hline
\textbf{Method} & \textbf{Primary Package Color} & \textbf{Package Shape} & \textbf{Package Material} & \textbf{Category} & \textbf{Weight (Acc@0.01)} & \textbf{Weight (Acc@0.05)} \\
\hline
Baseline (zero-shot) & 26.67 & 3.81 & 19.05 & 0.00 & 9.78 & 9.78 \\
Baseline w/ schema & 55.24 & 49.52 & 54.29 & 62.86 & 13.04 & 16.30 \\
ours w/o reasoning & 81.9 & 76.19 & 81.9 & 95.24 & 55.43 & 63.04 \\
ours w/o multi-turn & 73.33 & 75.24 & 79.05 & 89.52 & 54.35 & 69.57  \\
ours & \textbf{82.86} & \textbf{77.14} & \textbf{86.67} & \textbf{97.14} & \textbf{61.96} & \textbf{73.91} \\
\hline
\end{tabularx}
\end{table*}

\subsection{Cycle of Enrollment}

With the graph initialized and the schema in place, individual products can be processed and added to the knowledge graph iteratively. The Cycle of Enrollment consists of four key steps: \textbf{Extracting}, \textbf{Formatting and Inferring}, \textbf{Hierarchy Expansion}, and \textbf{Graph Pruning}. Additionally, while we primarily focus on studying KG construction from product images, our method inherently supports textual product description as input by skipping the \textbf{Extract} phase. Each phase addresses specific challenges in constructing a reliable and hierarchical product knowledge graph for e-commerce.

\textbf{Extracting} product descriptions from the raw image is the first step of enrolling an image. To tackle the challenge of extracting rich information from images, we employed a recent state-of-the-art open-source vision language model, InternVL2 \cite{https://arxiv.org/pdf/2312.14238, https://arxiv.org/pdf/2404.16821}. We first convert the generated schema into text descriptions to augment the original prompt. With schema description embedded in the user prompt, the VLM can generate unstructured or semi-structured text descriptions based on the input product image, ensuring that the generated descriptions cover all relevant product attributes. To maximize information coverage, we employ a multi-turn extraction process, where we prompt VLM to provide additional details in a second turn of the conversation. This will cover more visual cues compared to single-turn extraction to enable more accurate missing information inference in future steps. More turns are possible yet not employed for better speed and scalability.

\textbf{Formatting and Inferring} generated description comes after information extraction. While visible information is already included in the text description, the information is not directly usable. Similar to the extracting phase, we use multi-turn conversation to align the response with the predefined schema. Since some information may not be available from the image, we use Llama3.1-70B \cite{https://arxiv.org/pdf/2407.21783} to first analyze all the extracted text descriptions, encouraging intermediate reasoning steps \cite{https://arxiv.org/pdf/2201.11903}. Then based on the reasoning, we prompt the model to infer the remaining properties. After the generation in the first turn, we use SGLang \cite{https://arxiv.org/pdf/2312.07104} for regular expression constrained generation. The output is forced to be generated in JSON format, strictly following the data type and schema structure. This guarantees that the response will always be generated reliably containing all requested properties, and additionally ensures the response can be parsed programmatically.

\textbf{Hierarchical Expansion} attempts to introduce additional entities between the product node and the abstract category node. This phase is crucial for enhancing the knowledge graph's structure and utility in e-commerce applications. An LLM is prompted to analyze and generate an intermediate entity between a category property and the product name. This expansion is repeated several times so that multiple intermediate entities are inserted. For example, the category link starts with "Dark Chocolate Bar $\rightarrow$ Food and Beverage", an intermediate nodes "Chocolate" and "Dark Chocolate" are sequentially added during the expansion. The resulting link becomes "Dark Chocolate Bar $\rightarrow$ Dark Chocolate $\rightarrow$ Chocolate $\rightarrow$ Food and Beverage". To further improve the diversity of the graph, multiple hierarchical expansions are performed in parallel, each independently choosing intermediate nodes from the predicted tokens with top-k sampling. By introducing multiple levels of abstraction, the system can represent products at various levels of breadth with more fine-grain relationships.

\textbf{Pruning} is the final step of enrolling a product. When properties are created with LLM free-form generation, there can be entities sharing exactly the same or similar meaning, these can be merged into one node. In our method, we applied a simple yet effective method that merges all properties that share the same words in different order or letter cases. This reduces the complexity of the final knowledge graph so that downstream tasks can be performed more efficiently.

\begin{figure}[h]
  \centering
  \includegraphics[width=\linewidth]{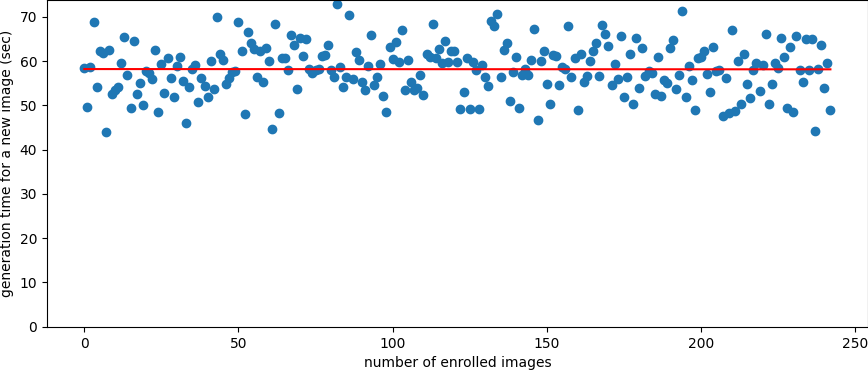}
    \Description{A plot of time taken in seconds against the number of enrolled images. The time taken to generate a knowledge graph scales linearly with the number of images.}
    \caption{ 
    Time taken to generate KG for an image remains similar as the number of images in the inventory increases.
    }
  \label{fig:fig2}
\end{figure}

This Cycle of Enrollment is executed for each product, allowing for the incremental growth of the knowledge graph and adapting to the rapid changes in the e-commerce domain. Because each product subgraph is generated independent of the size of the existing inventory, as shown in Figure \ref{fig:fig2}. The linear scaling allows our approach to be scaled to a large inventory size without increasing the cost of generating the graph for each image.

By combining advanced VLMs and LLMs with constrained generation, our approach can handle diverse product types and extract rich, consistent structured information from visual data, reliably following the schema. The resulting hierarchical knowledge graph provides a foundation for sophisticated e-commerce applications, including enhanced search capabilities, personalized recommendations, and advanced product analytics.

\subsection{Inventory Usage}
In the final stage after the enrollment, one can easily leverage the KG to perform user-defined tasks. Apart from the classic product description retrieval and recommendation systems, our method enables intelligent attribute inquiry like packaging materials and weight estimation, offering users a diverse range of applications.

\section{Experiment}
In this section, we first gather a product image dataset. Then, we use our method to generate a knowledge graph following the method described in the previous section. We compare the ground truth annotation with generation results and evaluate the effectiveness of our method based on several metrics. Unless otherwise specified, we use InternVL2-8B in bfloat16 and Llama3.1-70B in int4. The experiments are conducted on 6 RTX 4090 GPUs.

\subsection{Dataset Collection}
We collected 120 images and their corresponding metadata using BlueCart Walmart Data Product API\footnote{https://www.bluecartapi.com/}. The result contains information such as image, product name, and other information displayed on the Walmart website\footnote{https://www.walmart.com/}. Among these, 105 images are valid, we then manually labeled the properties Category, Primary Package Color, Package Material, Package Shape, and Weight based on our generated schema. We perform a series of experiments using only the images resized to 448x448 pixels as inputs.

\subsection{Results}
In this part, we will show our performance on the collected dataset compared to a baseline method. We set our baseline by modifying the zero-shot KG construction method proposed in AutoKG \cite{https://arxiv.org/pdf/2305.13168} to generate triples from an image using the product name as the subject. Following AutoKG, we use property names as the list of predicates to prompt InternVL2, and the objects of the generated triples are treated as the predicted properties. In addition to the zero-shot baseline, we add our schema-augmented prompt to the start of the baseline prompt, providing additional context for the expected response. Then, we perform an ablation study on our method. First, we evaluate the performance of our complete method. We then conduct ablation studies by removing the multi-turn conversation during VLM information \textbf{Extraction} or excluding the intermediate LLM reasoning step during the \textbf{Format and Infer} stage, to assess the impact of these components on the overall results. We evaluated the following predictions against the annotation. \textbf{Primary Package Color and Package Shape} are categorical properties that can be directly observed from the image with little reasoning and inference. \textbf{Package Material} is a categorical property that can be directly observed from the image, but requires some prior knowledge (e.g., material texture) to infer. \textbf{Category} is a categorical property that cannot be directly observed from the image, and requires inference with prior knowledge and contextual reasoning on information like brand name. \textbf{Weight} is a numerical property that can be found on most of the product images, but due to non-standardized units of measurement across products, calculations are needed to convert to the unit in schema (kg).

\begin{figure}[h]
  \centering
  \includegraphics[width=\linewidth]{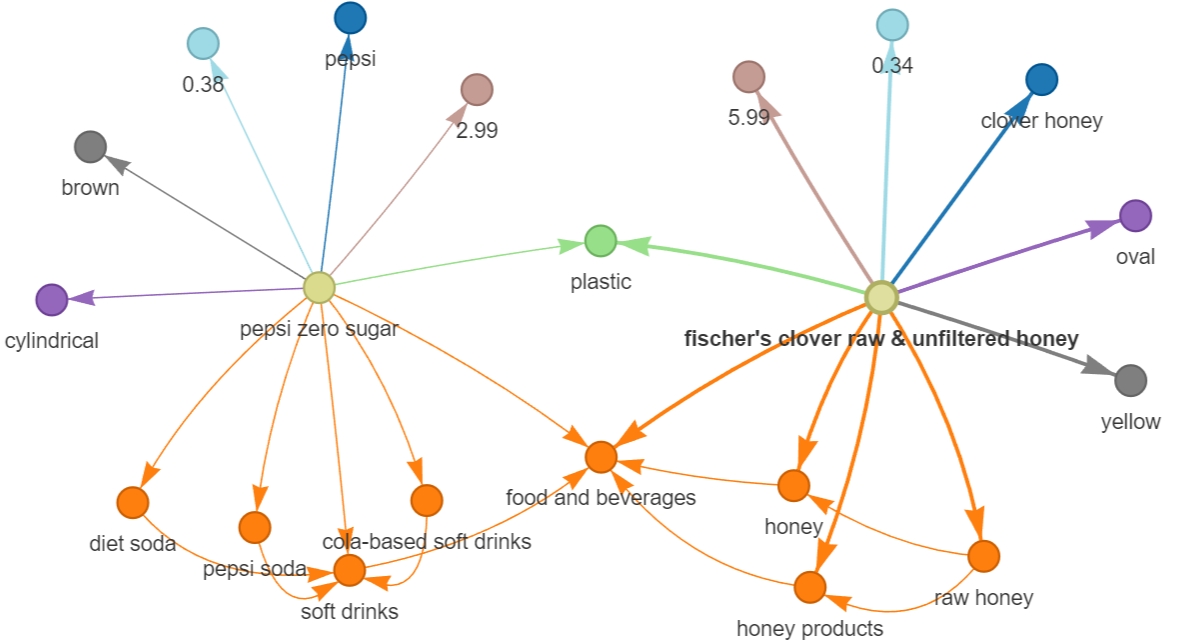}
    \Description{Example KG subgraph of 3 enrolled products.}
    \caption{ 
    Example KG subgraph of 3 enrolled products.
    }
  \label{fig:fig3}
\end{figure}

For categorical properties, accuracy is used as the metric. For numerical property, we first compute the error ratio $e$ between predicted value $v_{pred}$ and annotated value $v_{gt}$ by
\begin{equation}
    e = \frac{|v_{pred} - v_{gt}|}{v_{gt}}
\end{equation}
Then we use accuracy@threshold as our metrics, where a prediction is considered correct if the error ratio $e$ is strictly lower than the threshold. We report accuracy@0.01 and accuracy@0.05 for the numerical property. Table \ref{tab:results} shows our primary result against baseline and ablation studies. We also show a subgraph containing 3 enrolled products constructed our complete method in Figure \ref{fig:fig3}.

\subsection{Analysis}
As shown in Table \ref{tab:results}, our method exceeds directly prompting VLM for triple generation. By augmenting the baseline with our schema description, predictions for all categorical properties gain large improvements. We notice that this is mainly because when no schema descriptions are embedded in the prompt, VLM tends to give predictions that are not in the choices. Adding schema description provides contextual information for VLM to rectify its answers. 

By removing LLM reasoning from our method, performance on Weight prediction and Package Material drops significantly. Accuracy@0.05 for Weight dropped over 10\%. This shows that reasoning is important for analyzing more ambiguous properties that require contextual understanding. With weight information directly shown in most images, our method without reasoning fails to standardize units more frequently than the complete method. The LLM tends to directly provide weight in its original units of measurement, even though it is prompted to respond in kilograms. Our result shows that the reasoning step leverages the LLM's ability to incorporate external knowledge and perform context-sensitive analysis, which is crucial for property inference.

Furthermore, the drop in performance when removing the multi-turn conversation during VLM information extraction highlights the importance of including diverse additional visual information in image descriptions. The multi-turn process allows the model to progressively add visual cues and details into the descriptions, which could be beneficial for subsequent steps to analyze and dynamically adjust predicted properties based on the context.

Even without reasoning or multi-turn conversation, our method still outperforms the baseline by a large margin, showing the robustness of our method when constructing links from image data.

\section{Limitations}
While our work shows promising results on various metrics using high-quality images, additional work may be required for low-resolution images.

\section{Conclusion}
In this paper, we propose a novel method that fully automatically generates a knowledge graph from scratch using only image data. We propose several collaborative components to analyze and infer schema-compliant properties from each product image. We propose a benchmark for knowledge graph generation from images, with emphasis on the correctness of generated properties. We compare our method against an adaptation of a previous work \cite{https://arxiv.org/pdf/2305.13168}, and perform ablation studies, showing the effectiveness of our method and several key features.


\bibliographystyle{ACM-Reference-Format}
\bibliography{main.bib}

\end{document}